\documentclass[runningheads]{llncs}

\usepackage[utf8]{inputenc} 
\usepackage[T1]{fontenc}    
\usepackage{hyperref}       
\usepackage{url}            
\usepackage{booktabs}       
\usepackage{amsfonts}       
\usepackage{nicefrac}       
\usepackage{microtype}      
\usepackage{lipsum}
\usepackage{fancyhdr}       
\usepackage{graphicx}       
\usepackage{marvosym}
\graphicspath{{figs/}}     
\usepackage{geometry}
\title{CIPER: Combining Invariant and Equivariant Representations Using Contrastive and Predictive Learning}
\titlerunning{CIPER}

\author{
  Xia Xu\inst{1,2} \and Jochen Triesch\inst{1}\textsuperscript{\Letter}
}
\institute{Frankfurt Institute for Advanced Studies, Ruth-Moufang-Straße 1, Frankfurt am Main, Germany \and Xidian-FIAS International Joint Research Center, Technology Road 9, Xi'an, China
\email{\{xiaxu,triesch\}@fias.uni-frankfurt.de}}

\begin{document}
\maketitle

\begin{abstract}
Self-supervised representation learning (SSRL) methods have shown great success in computer vision. In recent studies, augmentation-based contrastive learning methods have been proposed for learning representations that are invariant or equivariant to pre-defined data augmentation operations. However, invariant or equivariant features favor only specific downstream tasks depending on the augmentations chosen. They may result in poor performance when the learned representation does not match task requirements.
Here, we consider an active observer that can manipulate views of an object and has knowledge of the action(s) that generated each view. We introduce Contrastive Invariant and Predictive Equivariant Representation learning (CIPER). CIPER comprises both invariant and equivariant learning objectives using one shared encoder and two different output heads on top of the encoder. One output head is a projection head with a state-of-the-art contrastive objective to encourage invariance to augmentations. The other is a prediction head estimating the augmentation parameters, capturing equivariant features. Both heads are discarded after training and only the encoder is used for downstream tasks. We evaluate our method on static image tasks and time-augmented image datasets. Our results show that CIPER outperforms a baseline contrastive method on various tasks. Interestingly, CIPER encourages the formation of hierarchically structured representations where different views of an object become systematically organized in the latent representation space.
\end{abstract}


\begin{figure}[t]
\centering
\includegraphics[width=0.75\columnwidth]{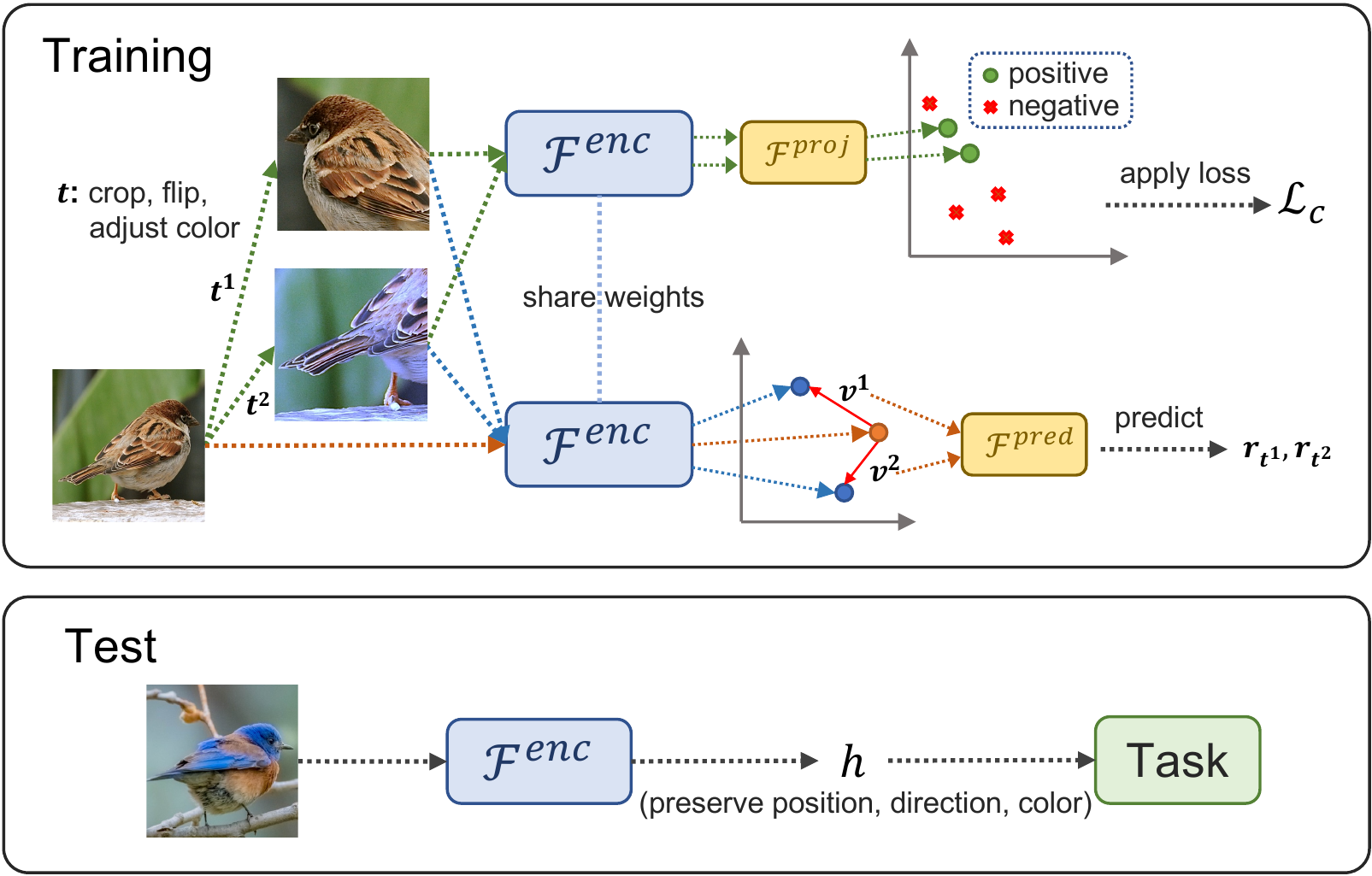}
\caption{Overview of CIPER. {\bf Top:} During training, $t^{1}$ and $t^{2}$ are two randomly sampled augmentations. $\mathcal{F}^{\rm proj}$ and $\mathcal{F}^{\rm pred}$ are the contrastive and predictive projectors applied on the same encoder $\mathcal{F}^{\rm enc}$. The InfoNCE contrastive loss $\mathcal{L}_{c}$ is applied to the output of the projector $\mathcal{F}^{\rm proj}$. The prediction head $\mathcal{F}^{\rm pred}$ predicts the parameters of augmentations $r_{t^1}$ and $r_{t^2}$ given the difference between the representations of the anchor image and the augmented images in the latent space ($v^{1}$ and $v^{2}$). {\bf Bottom:} During the test phase, the projectors are discarded and the output of the encoder $h$ is used for downstream tasks. Images are retrieved from \cite{tadepalli_english_2021,italia_english_2009} under CC BY-SA 4.0 licence.}
\end{figure}

\section{Introduction}
Recent advances in self-supervised representation learning (SSRL) have achieved performance comparable to supervised methods\cite{liu2021self}. SSRL utilizes internal structures of the data as supervisory information. Among SSRL methods, contrastive learning with deep neural networks has shown promising performance\cite{chen2020simple,laskin2020curl,radford2021learning}. Recent progress in contrastive learning has been achieved through strong image augmentations such as random cropping, color jittering, and random horizontal flipping, assuming these operations do not alter the semantic meaning of an image\cite{chen2020simple}. Augmentations encourage invariance, and representations of differently augmented versions of the same image are made similar, while representations of different images are made dissimilar to avoid representation collapse. However, strong augmentations impose a bias that downstream tasks require little augmentation-related information in the representation. Otherwise, contrastive learning would lose valuable information\cite{von2021self}. For example, when the task is to recognize hand-written digits, while the model learns to be invariant to in-plane image rotations, it will be incapable of distinguishing \textquotedblleft 9\textquotedblright{} from \textquotedblleft 6\textquotedblright.

On the other hand, equivariant representation learning aims to learn features that change according to augmentations. This learning paradigm has been facilitated using contrastive and predictive learning paradigms for static images and time series\cite{gidaris2018unsupervised,gidaris2019boosting,zhang2019aet,dangovski2021equivariant,feng2019self,jenni2021time}. Equivariant representations, like their counterpart, may focus on certain features and ignore other useful information. In the context of self-driving cars, for instance, representations trained to be equivariant to image rotations may focus on the road's direction and ignore cars in images. These representations may be useful for road direction detection, but they may be of little use for, e.g., identifying cars on roads in the context of some other task.

To balance the trade-off between invariant and equivariant features, some recent studies have proposed to combine them\cite{zhang2020equivariance,dangovski2021equivariant,rizve2021exploring,wang2021residual,xiao2020should}. Here, we propose a new approach, CIPER(Contrastive Invariant and Predictive Equivariant Representation learning), which trains an encoder with two output heads: one for contrastive learning and the other for predictive learning on augmentation parameters. This method enables the extraction of both invariant semantic meanings and the factors manipulated by the augmentations. By doing so, CIPER maximizes the use of augmentations without biasing the learning process exclusively towards either invariant or equivariant features. The resultant representations become hierarchically organized: views of the same object become organized into clusters as for contrastive learning methods. Moreover, in CIPER these clusters develop a systematic substructure reflecting augmentation-related information.

\section{Related Work}
\subsection{Invariant representation learning.}

Contrastive learning is a simple and effective method to promote invariance to augmentations. InfoNCE\cite{oord2018representation} uses a unified categorical cross-entropy loss to identify positive pairs against negative ones. SimCLR\cite{chen2020simple} improves generalization by using heavy data augmentations and large batch sizes. Several other methods also bring the representations of two augmented versions of an image together\cite{he2020momentum}. ReLIC\cite{mitrovic2020representation} introduces a regularizer to isolate style information. A simulated dataset is used in \cite{von2021self} to control the separation of content and style. Our method extends the conventional contrastive learning by preserving augmentation-related information.

\subsection{Equivariant representation learning.}

Inspired by the equivariant map in mathematics, representation learning with deep neural networks seeks to create equivariant representations that reflect the applied transformations\cite{pmlr-v48-cohenc16}. Equivariant representation paradigms aim to create representation spaces that are mathematically equivariant to group actions (image transformations)\cite{pmlr-v48-cohenc16}. Other methods aim to create representations that are equivariant to augmentation parameters such as image transformation matrices\cite{feng2019self,zhang2019aet}, pre-defined tasks like four-fold rotation prediction\cite{gidaris2018unsupervised}, and temporal auxiliary tasks for time series\cite{jenni2021time}. Prelax\cite{wang2021residual} uses residual relaxation and additional augmentations to encourage equivariance beyond invariance. CIPER predicts all augmentation parameters given the difference vector between the original image representation and the augmented one, unlike Prelax, which requires a target network, extra augmentations, and residual relaxation. CIPER adopts the SimCLR network structure and augmentations without any additional modifications.

\subsection{Combining invariant and equivariant learning.}

Several methods have been proposed to combine invariant and equivariant representations. Closest to our work are \cite{zhang2020equivariance,dangovski2021equivariant,rizve2021exploring,wang2021residual,xiao2020should}. Specifically, LooC\cite{xiao2020should} learns subspaces invariant to all but one augmentation, which may limit extension to many augmentations. E-SSL\cite{dangovski2021equivariant} uses an extra prediction head to predict an augmentation parameter, arguing that representations should be either invariant or equivariant to a specific augmentation. However, we show that both can be encoded at the same time as clusters with organized substructures. In addition, we do not pick specific augmentations other than those already used in SimCLR while E-SSL chooses a four-fold rotation augmentation for the equivariant objective which requires training on 4 extra augmentated versions of an image, resulting in a huge training burden. A residual relaxation-based method\cite{wang2021residual} has also been proposed. It uses extra augmentations to encourage equivariance beyond invariance for few-shot learning problems. Our method combines contrastive and predictive objectives to naturally encode both invariant and equivariant information without explicit relaxation. Furthermore, although these methods manage to encode both, it is still unclear how equivariant information is encoded in the representations, while CIPER manages to incorporate this as substructures within a clustered representation. 

\section{Methods\label{sec:Methodology}}

Given unlabelled high-dimensional data $x\in\mathbb{R}^{m}$, self-supervised representation learning trains an encoder network $\mathcal{F}^{\rm enc}$ to extract low-dimensional representations $h = \mathcal{F}^{\rm enc}(x) \in\mathbb{R}^{n}, n \ll m$, that are informative for a downstream task target variable $g$. The goal is to maximize $\mathcal{I}\left(h;g\right)$. Augmentations are general transformations $t\in\mathcal{T}$ that create augmented views $x^{t}$ from raw data $x$ as proxies for downstream tasks. Typically, a single augmentation $t$ is a sequence of random image operations parameterized by $r_{t}$. For example, if $t$ is random cropping at $\left(i,j\right)$ and random horizontal flipping with an indicator $p$, then $r_{t}=\left[i,j,p\right]$. For some datasets we also consider time-based augmentations, where an active observer can change the viewpoint of an object and/or view it against a different background \cite{aubret2022time}. In the following, we refer to these as {\em dataset-provided augmentations} (see below). Throughout the paper, augmentations are performed in a fixed order (see Sec.~\ref{subsec:datasets}).

\subsection{Invariant Contrastive Learning.\label{subsec:Invariant-Contrastive-Learning}}

Contrastive learning generates representations that are invariant to data augmentations. For an anchor data point $x_{i}$, its two differently augmented versions $x_{i}^{t_{i}^{1}}$ and $x_{i}^{t_{i}^{2}}$ are positive samples, and different data points with different augmentations $x_{j}^{t_{j}^{1}}$ and $x_{j}^{t_{j}^{2}}$ are negative samples. The goal is to maximize the similarities between positive pairs and minimize the similarities between positive and negative samples. CIPER adopts SimCLR, in which a batch of $N$ data points are augmented twice and encoded by $\mathcal{F}^{\rm enc}$ to obtain representations $h_{i}$. The representations are projected by $\mathcal{F}^{\rm proj}$ to $z$ where the InfoNCE loss $\mathcal{L}_{c}$ is applied. If the training fully converges, which is rare in practice, then $z_{i}^{t_{i}^{1}}$, $z_{i}^{t_{i}^{2}}$ and $z_{i}$ should be indistinguishable. Similarly, $h$ should also reflect little information about the augmentations. We show in Sec.~\ref{sec:results} that some augmentation-related information remains in the latent representation despite the contrastive learning objective. Furthermore, reducing $\mathcal{I}\left(\mathcal{F}^{\rm enc}\left(x^{t}\right);t\right)$ is not guaranteed to increase $\mathcal{I}\left(\mathcal{F}^{\rm enc}\left(x^{t}\right);g\right)$.

\subsection{Equivariant Predictive Learning.}

We consider equivariant representations that change with input data according to an equivariant map from mathematics. The representation reflects how image augmentations alter data rather than preserving input information. Equivariant representation learning maximizes $\mathcal{I}\left(\mathcal{F}^{\rm enc}\left(x^{t}\right); t\right)$ where $\mathcal{F}^{\rm enc}$ is an encoder with a prediction head and loss as in Sec.~\ref{subsec:Invariant-Contrastive-Learning}. This may increase $\mathcal{I}\left(\mathcal{F}^{\rm enc}\left(x^{t}\right); g\right)$ if $t$ affects task $g$.
We use static image augmentations from SimCLR (Tab.~\ref{table:augmentations}) with 10 dimensions and dataset-specific special augmentations (Sec.~\ref{subsec:datasets}). In CIPER, we add a predictor $\mathcal{F}^{\rm pred}$ to $\mathcal{F}^{\rm enc}$ to predict augmentation parameters, thereby maximizing $\mathcal{I}\left(\mathcal{F}^{\rm enc}\left(x^{t}\right); r_{t}\right)$. This is straightforward and easy to implement. Following Sec.~\ref{subsec:Invariant-Contrastive-Learning}, we augment each image twice as a pair. For an anchor image $x_{i}$ and its augmentation $x_{i}^{t_{i}^{1}}$, the prediction head $\mathcal{F}^{\rm pred}$ takes their representation difference $h_{i}-h_{i}^{t_{i}^{1}}$ as input and feeds it into the prediction objective. For static augmentations, we normalize the target (the parameters) across batches and use the mean squared error (MSE) loss:
\begin{equation}
\mathcal{L}_{p}=-\frac{\sum_{i}^{N}\left(\sum_{k}^{M}\left(r^{t_{i}^{1}}_{k}-\hat{r}^{t_{i}^{1}}_{k}\right)^{2}+\sum_{k}^{M}\left(r^{t_{i}^{2}}_{k}-\hat{r}^{t_{i}^{2}}_{k}\right)^{2}\right)}{2\cdot N\cdot M} \, ,
\label{eq:prediction-loss}
\end{equation} where $\hat{r}$ is the output of the prediction head, $M$ is the total number of dimensions of the augmentations and $N$ is the batch size. For dataset-provided augmentations, we use cross-entropy for categorical and MSE for continuous parameters.

\begin{table*}[tb]
\caption{Augmentations and their parameters. $p$ indicates the probability of applying a particular augmentation. Binary parameters are indicators for whether an image is augmented. Viewpoint and Session are dataset-specific augmentations (See Sec.~\ref{subsec:datasets}).Parameters are randomly drawn and returned using the TorchVision package\cite{marcel2010torchvision}.}\label{table:augmentations}
\resizebox{\textwidth}{!}{
\begin{tabular}{|c|c|c|c|c|}
\hline 
Augmentation & Type & \# Dims & Meaning & Setting\\
\hline 
\hline 
Cropping & continuous & 4 & {[}location x, location y, height, width{]} & $p=1$; scale: (0.2, 1.0)\\
\hline 
Horizontal Flip & binary & 1 & 0 for not flipped, 1 otherwise & $p=0.8$\\
\hline 
Color Jittering & continuous & 4 & {[}brightness, contrast, saturation, hue{]} & $p=0.8$\\
\hline 
Grayscale & binary & 1 & 0 for grayscale, 1 otherwise & $p=0.2$\\
\hline 
Viewpoint(TDW only) & continuous & 3 & change in azimuth, elevation, distance &  $p=1$; normalized\\
\hline
Session(CORe50 only) & categorical & 1 & target session & $p=1$\\
\hline
\end{tabular}}
\end{table*}

\subsection{Combining Invariant and Equivariant Representations.}

To allow the combination of invariant and equivariant learning, we simply apply at the same time the InfoNCE loss on the output $z$ of the prediction head and the prediction loss (Eq.~\ref{eq:prediction-loss}) on the output $\hat{r}$ of the prediction head. The total loss is then:
\begin{equation}
\mathcal{L}=\mathcal{L}_{c}+\alpha\cdot\mathcal{L}_{p} \, ,
\end{equation}
where $\alpha$ is the weighting hyper-parameter of the predictive loss. We use $\alpha=1.0$ for CIPER except for the ablation study described below. After training, we discard both heads and use only the encoder for $h$. Combining two counteracting objectives may seem counter-intuitive, but Sec.~\ref{sec:results} shows that $h$ benefits from both. 

\subsection{Datasets, Augmentations, Experimental Setup. \label{subsec:datasets}}

We evaluate our method on CIFAR10\cite{krizhevsky2009learning} and two image datasets with time-based augmentations: CORe50\cite{lomanco2017core50} and TDW\cite{schneider2021contrastive}. CORe50 contains videos of objects (e.g., cups, balls) in different environments (e.g., kitchen, garden), called sessions. We use 2 sessions for training and 9 for testing to avoid trivial session encoding. TDW consists of sequences of rendered objects from different perspectives. We follow the split in \cite{schneider2021contrastive}. A common augmentation\cite{schneider2021contrastive} is sampling the next image in the sequence as the positive pair for contrastive learning. However, this type of augmentation lacks the ability to be parameterized. We propose new augmentations using the view and session parameters of TDW and CORe50, respectively. In TDW, we manipulate the 3D view (azimuth, elevation, distance) of an object. In CORe50, we use 11 session categories (2 for training). An object is augmented by randomly changing the session and the target session label becomes the augmentation parameter. We call these ``Dataset Aug.'' in Tab.~\ref{table:CORe50-TDW}. We also apply standard image transformations (``Image Aug.'') from SimCLR (see Tab.~\ref{table:augmentations}). We compare these augmentations on CIFAR10, CORe50 and TDW.

We use ResNet18\cite{he2016deep} as our encoder backbone with a 3$\times$3 convolutional kernel and no max pooling in the first layer, as in SimCLR\cite{chen2020simple}. The projection head is a two-layered MLP\cite{rumelhart1985learning} with 2048 hidden units in each layer and batch normalization\cite{ioffe2015batch} and ReLU\cite{nair2010rectified} after each layer, following E-SSL\cite{dangovski2021equivariant}. The predictor is another two-layered MLP with 512 hidden units and LayerNorm\cite{ba2016layer} and ReLU after the first layer. The output dimension depends on the augmentation parameters to be predicted. See Tab.~\ref{table:augmentations} for more details on the augmentations. For standard image augmentations, we use TorchVision\cite{marcel2010torchvision} with a wrapper to apply random augmentations and access their parameters. To facilitate easier encoding of relative position information for the encoder, we concatenate Cartesian coordinates with RGB-channels for positional encoding as in \cite{murase2020can}. We use stochastic gradient descent (SGD) with 0.03 initial learning rate (cosine decay), $5\times10^{-4}$ weight decay and 0.9 momentum. We train both networks for 800/100 epochs with 256/64 batch size on CIFAR10/TDW and CoRE50 using Sec.~\ref{sec:Methodology} losses. We freeze the encoder and train a linear layer on top of it for linear evaluation as in SimCLR. For CIFAR10, we use training data and test accuracy. For TDW and CORe50, we also train a regressor/classifier for view/session identification. For CORe50, we sample from 9 unseen sessions for both training and test sets. The linear layer is trained for 100 epochs with SGD ($10^{-6}$ weight decay, 0.9 Nesterov momentum\cite{sutskever2013importance,nesterov1983amf}) and initial learning rate 1 (decayed by 3.33 every 10 epochs). We report mean accuracy and standard deviations over five runs with different random seeds.

\section{Results\label{sec:results}}
\begin{table*}[tb]
\caption{CORe50 and TDW Results. The object and session classification accuracy (\%) are referred to as Obj Acc. and Sess Acc., respectively. For the TDW dataset, we mark the regression coefficient of determination of the view as View $R^2$. Regular image-based augmentations and dataset-specific augmentations are marked as Image Aug. and Data Aug. Each reported value is shown as mean ± std across 5 runs.}\label{table:CORe50-TDW}
\resizebox{\textwidth}{!}{
\begin{tabular}{|c|c|c|c|c|}
\hline 
Method & CORe50 (Obj Acc.) & TDW (Obj Acc.) & CORe50 (Sess Acc.) & TDW (View $R^2$)\\
\hline 
Random encoder & 18.44 ± 10.01 & 60.73 ± 8.39 & 81.29 ± 5.51 & 0.11 ± 0.06 \\
\hline 
Contrastive w/ Image Aug. & 66.95 ± 0.84 & 95.91 ± 0.17 & 99.08 ± 0.09 & 0.21 ± 0.01 \\
\hline 
Contrastive w/ Dataset Aug. & 60.55 ± 0.87 & 98.32 ± 0.15 & 96.81 ± 0.61 & 0.13 ± 0.07 \\
\hline 
E-SSL w/ Image Aug. & 59.59 ± 0.68 & 96.24 ± 0.29 & 98.86 ± 0.23 & 0.29 ± 0.03 \\
\hline
CIPER w/ Image Aug. & {\bf 75.43 ± 0.61} & 97.36 ± 0.16 & {\bf 99.97 ± 0.02 } & 0.52 ± 0.01 \\
\hline 
CIPER w/ Dataset Aug. & 67.46 ± 0.41 &  {\bf 98.92 ± 0.41} & 99.72 ± 0.07 & {\bf 0.87 ± 0.02}\\
\hline 

\end{tabular}}
\end{table*}

 We evaluate CIPER on the CORe50 and TDW datasets, which offer additional augmentation choices. We also report the results of classifying the session in CORe50 and regression on the view parameters of TDW datasets. From the results shown in Tab.~\ref{table:CORe50-TDW} we observe that dataset-specific augmentations can achieve better performance than general image-based augmentations. We further find that the contrastive representations would fail to achieve high view identification $R^2$ score, while CIPER can solve this problem and improve the contrastive method on other tasks. These findings show that the combined CIPER objective retains rich information about the augmentations compared to the invariant objective alone. Another interesting finding is that compared with the untrained randomly-initialized encoder, the contrastive representations achieve higher session classification accuracy on CORe50. Furthermore, we evaluate our encoder on CIFAR10 with the contrastive objective and with CIPER, which uses both contrastive and predictive objectives (Tab.~\ref{table:CIFAR10-result}). The encoder trained with the predictive network alone can not match the performance of the state-of-the-art contrastive learning methods. This shows that CIPER is robust to situations where one of the two objectives alone would fail. The results support our hypothesis that the invariant representations generally discard the information about the augmentation and the equivariant representations do the opposite, while neither of them achieves the goal of discarding or preserving augmentation information completely. However, by combining both contrastive and predictive objectives CIPER can robustly preserve rich information and achieves better performance in both situations.

 \begin{table*}[tb]
\caption{Linear classification accuracy (\%) on CIFAR10. The ``(n$\times$)'' means that n augmented/original images are used for one image sample during training. For a fair comparison, only the methods with the SimCLR backbone are shown. The results of SimCLR and E-SSL are retrieved from \cite{dangovski2021equivariant}.}\label{table:CIFAR10-result}
\begin{centering}
\resizebox{0.4\textwidth}{!}{
\begin{tabular}{|c|c|}
\hline 
Method & Acc.\\
\hline 
\hline 
SimCLR (2$\times$) & 91.1\\
\hline 
SimCLR (re-production) (2$\times$) & 91.7 ± 0.1\\
\hline 
E-SSL (6$\times$) & 94.1 ± 0.0\\
\hline
Predictive (3$\times$) & 53.4 ± 2.3\\
\hline
CIPER (3$\times$) & 92.2 ± 0.3\\
\hline 
\end{tabular}}
\par\end{centering}
\end{table*}

\begin{figure*}[tb]
\begin{centering}
\includegraphics[width=0.7\columnwidth]{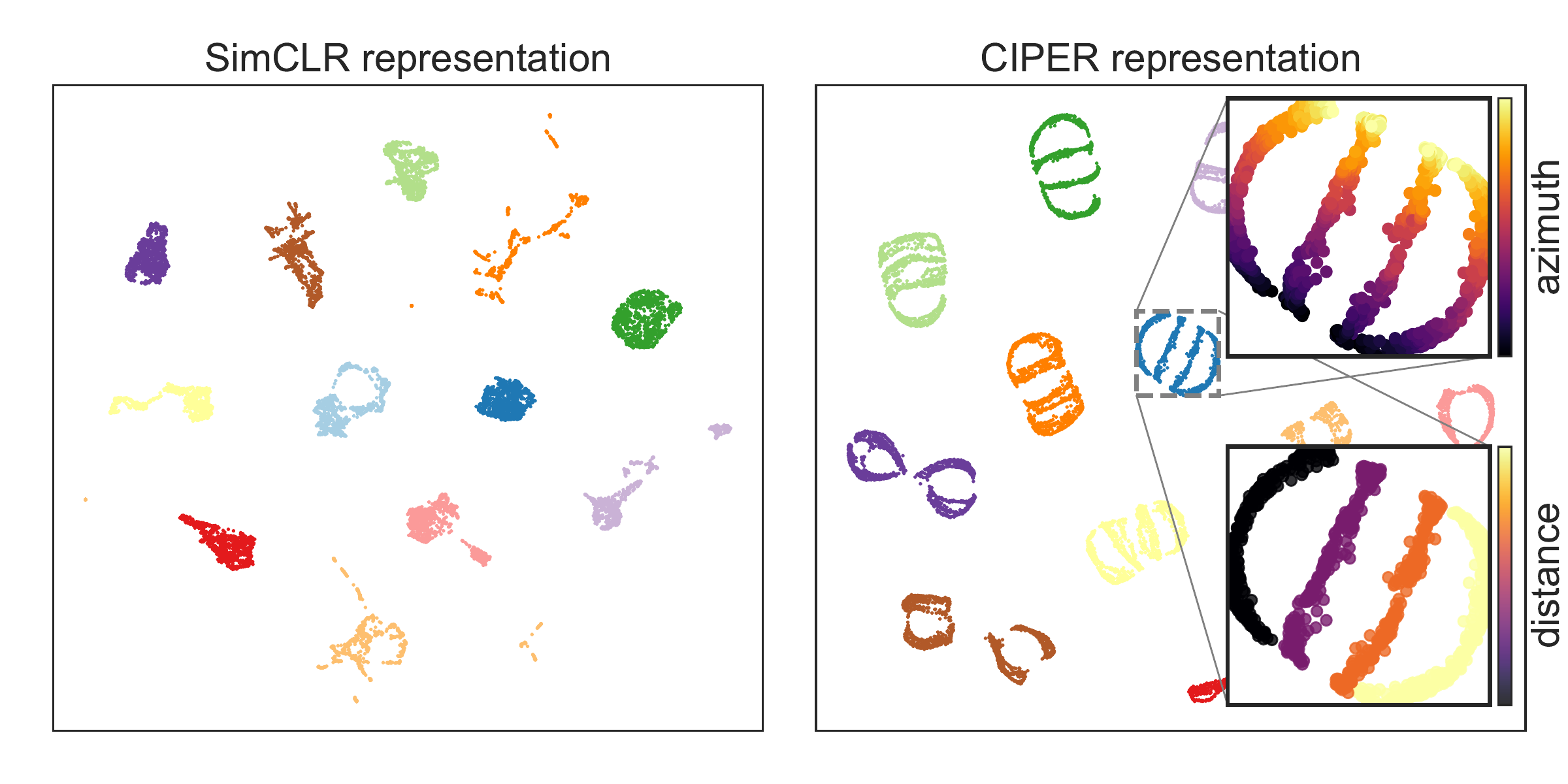}\includegraphics[width=0.18\columnwidth]{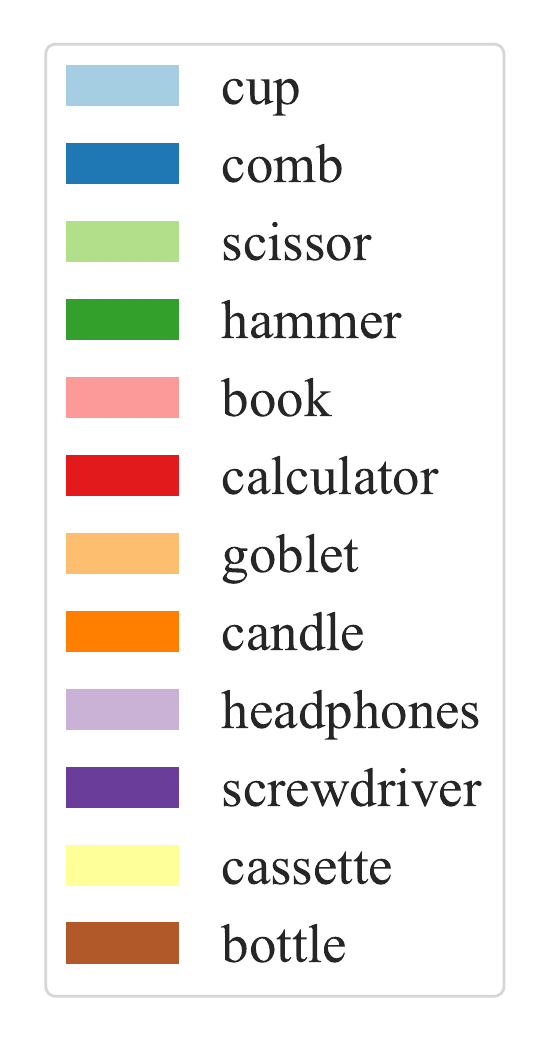}
\caption{PacMap\cite{wang2021understanding} representation visualizations of the TDW dataset after 100 epochs of training with dataset-specifc augmentations. {\bf Left:} representations with SimCLR objective. {\bf Right:} representations with CIPER objective. Note how CIPER encodes and systematically organizes view point information without losing the ability to cluster representations of the same object.}
\label{fig:representation-visualization}
\end{centering}
\end{figure*}

In Fig.~\ref{fig:representation-visualization} we visualize the learned representations for the TDW dataset. We use PacMap\cite{wang2021understanding} to reduce the dimension of the representations to 2, revealing a clustering of learned representations with the contrastive and the full CIPER objective. We can see that the contrastive objective drives representations to form compact clusters of objects, while the CIPER objective also encodes structure related to the augmentations.

\begin{figure*}[t]
\centering
\includegraphics[width=0.65\columnwidth]{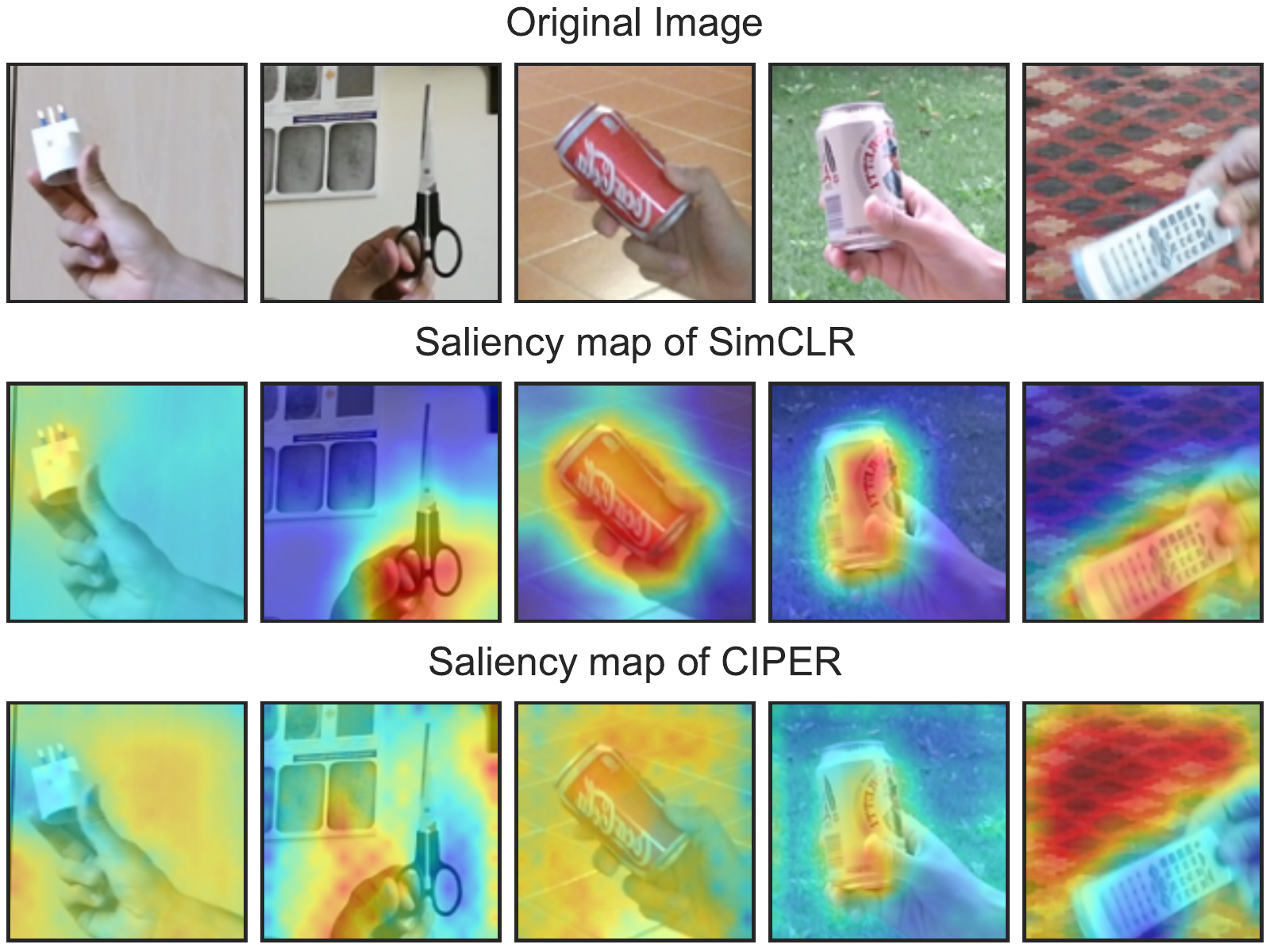}
\caption{Example FullGrad\cite{srinivas2019full} saliency maps of images from the CORe50 dataset for the encoders trained with SimCLR and CIPER (ours). The augmentations used are the data-specific augmentation defined in Sec.~\ref{subsec:datasets}. As the prediction head seeks to identify the recording session in CORe50, CIPER also pays attention to the background.}
\label{fig:saliency-map}
\end{figure*}

To further illustrate the representations learned by CIPER, we adopt the FullGrad method\cite{srinivas2019full} on CORe50 using the dataset-specific augmentation and the linear object classifier obtained at test phase to generate saliency maps as shown in Fig.~\ref{fig:saliency-map}. Since CIPER retains information about the object as well as the augmentation (in this case: against which background the object is seen) the saliency maps cover both object and background.
This finding suggests that CIPER could be utilized to learn disentangled representations of objects against backgrounds.

\begin{figure*}[t]
\begin{centering}
\includegraphics[width=0.9\columnwidth]{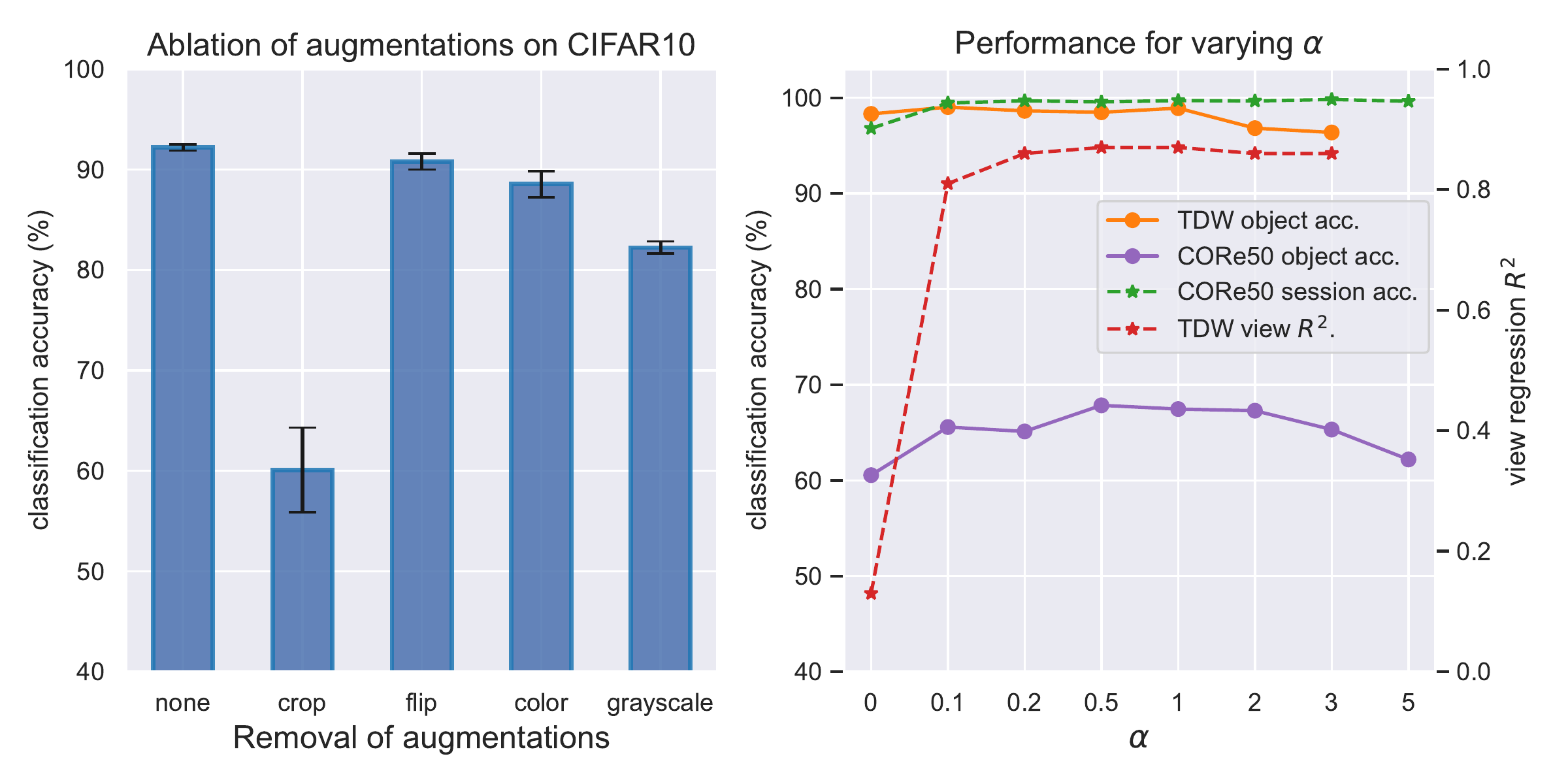}
\caption{Ablation studies. {\bf Left:} CIFAR10 object classification accuracy for ablation of single augmentations. {\bf Right:} Effect of varying $\alpha$. Object/Session classification accuracy on TDW and CORe50 (solid) and view regression $R^2$ score on TDW (dotted). Training with $\alpha > 3$ is unstable on TDW.}
\label{fig:ablation}
\end{centering}
\end{figure*}

To study what augmentations contribute most to the downstream classification task when training with CIPER, we remove each augmentation and test the performance. Similar to the findings in \cite{chen2020simple}, our results in Fig.~\ref{fig:ablation} (left) show that every augmentation benefits the downstream task while random cropping and color jittering contribute most to the performance.

Next, to study the impact of the prediction head, we vary the weighting hyper-parameter $\alpha$ and conduct experiments on TDW and CORe50. Results are shown in Fig.~\ref{fig:ablation} (right). Intermediate values of $\alpha$ lead to highest object classification accuracy on both TDW and CORe50, as well as higher session classification accuracy and lower view regression $R^2$ score on CORe50 and TDW, respectively. Based on this, we set $\alpha=1$ as default for other experiments.

\section{Discussion}

We have proposed a new method (CIPER) for combining invariant and equivariant objectives in self-supervised representation learning. Invariance is promoted via the popular SimCLR method. For equivariance we adopted a prediction approach that learns to estimate the parameters of the augmentations applied to the input data. In the test phase, the output heads are simply discarded and the encoder is used for downstream tasks.

Experiments show the benefits of our method compared to similar state-of-the-art works. In particular, we show that the incorporation of the equivariant objective of CIPER improves the representation. 

The use of the TDW and CORe50 datasets allowed us to consider an {\em active observer} that can manipulate the view of an object, since the used augmentations were essentially view point (TDW) and background (CORe50) changes. The incorporation of the predictive objective is therefore related to the difference between a passive observer that only sees different views of an object and an active observer that has access to the information {\em how} the view changed between two positive samples. CIPER exploits this additional information to learn an improved representation. This suggests that learning in biological vision systems may also benefit from active control over the viewpoint by an active observer. In fact, the combination of invariant and equivariant representation learning objectives used in CIPER is reminiscent of the separation of the primate visual system into a so-called ventral or ``what'' stream for invariant object recognition and a so-called dorsal or ``where/how'' stream for physical interaction with objects. This may reflect a general design principle for versatile vision systems capable of supporting qualitatively different tasks.

\section{Acknowledgements}

This research was supported by the research group ARENA (Abstract Representations in Neural Architectures) of the Deutsche Forschungsgemeinschaft (DFG)  under grant agreement TR 881/10-1. We also acknowledge support by ``The Adaptive Mind'' and ``The Third Wave of Artificial Intelligence,'' funded by the Excellence Program of the Hessian Ministry of Higher Education, Science, Research and Art. JT is supported by the Johanna Quandt foundation.

\bibliographystyle{splncs04}
\bibliography{references}  

\begin{thebibliography}{10}
\providecommand{\url}[1]{\texttt{#1}}
\providecommand{\urlprefix}{URL }
\providecommand{\doi}[1]{https://doi.org/#1}

\bibitem{aubret2022time}
Aubret, A., Ernst, M., Teuli{\`e}re, C., Triesch, J.: Time to augment
  contrastive learning. arXiv preprint arXiv:2207.13492  (2022)

\bibitem{chen2020simple}
Chen, T., Kornblith, S., Norouzi, M., Hinton, G.: A simple framework for
  contrastive learning of visual representations. In: International conference
  on machine learning. pp. 1597--1607. PMLR (2020)

\bibitem{pmlr-v48-cohenc16}
Cohen, T., Welling, M.: Group equivariant convolutional networks. In: Balcan,
  M.F., Weinberger, K.Q. (eds.) Proceedings of The 33rd International
  Conference on Machine Learning. Proceedings of Machine Learning Research,
  vol.~48, pp. 2990--2999. PMLR, New York, New York, USA (20--22 Jun 2016)

\bibitem{dangovski2021equivariant}
{Dangovski}, R., {Jing}, L., {Loh}, C., {Han}, S., {Srivastava}, A., {Cheung},
  B., {Agrawal}, P., {Solja{\v{c}}i{\'c}}, M.: {Equivariant Contrastive
  Learning}. arXiv e-prints arXiv:2111.00899 (Oct 2021)

\bibitem{feng2019self}
Feng, Z., Xu, C., Tao, D.: Self-supervised representation learning by rotation
  feature decoupling. In: Proceedings of the IEEE/CVF Conference on Computer
  Vision and Pattern Recognition. pp. 10364--10374 (2019)

\bibitem{gidaris2019boosting}
Gidaris, S., Bursuc, A., Komodakis, N., P{\'e}rez, P., Cord, M.: Boosting
  few-shot visual learning with self-supervision. In: Proceedings of the
  IEEE/CVF international conference on computer vision. pp. 8059--8068 (2019)

\bibitem{gidaris2018unsupervised}
{Gidaris}, S., {Singh}, P., {Komodakis}, N.: {Unsupervised Representation
  Learning by Predicting Image Rotations}. arXiv e-prints arXiv:1803.07728 (Mar
  2018)

\bibitem{he2020momentum}
He, K., Fan, H., Wu, Y., Xie, S., Girshick, R.: Momentum contrast for
  unsupervised visual representation learning. In: Proceedings of the IEEE/CVF
  conference on computer vision and pattern recognition. pp. 9729--9738 (2020)

\bibitem{he2016deep}
He, K., Zhang, X., Ren, S., Sun, J.: Deep residual learning for image
  recognition. In: Proceedings of the IEEE conference on computer vision and
  pattern recognition. pp. 770--778 (2016)

\bibitem{ioffe2015batch}
Ioffe, S., Szegedy, C.: Batch normalization: Accelerating deep network training
  by reducing internal covariate shift. In: International conference on machine
  learning. pp. 448--456. PMLR (2015)

\bibitem{italia_english_2009}
Italia, C.G.f.R.: English: {Italian} {Sparrow} pair (Jan 2009)

\bibitem{jenni2021time}
Jenni, S., Jin, H.: Time-equivariant contrastive video representation learning.
  In: Proceedings of the IEEE/CVF International Conference on Computer Vision.
  pp. 9970--9980 (2021)

\bibitem{krizhevsky2009learning}
Krizhevsky, A.: Learning multiple layers of features from tiny images. Tech.
  rep., Computer Science, University of Toronto (2009)

\bibitem{von2021self}
von K{\"u}gelgen, J., Sharma, Y., Gresele, L., Brendel, W., Sch{\"o}lkopf, B.,
  Besserve, M., Locatello, F.: Self-supervised learning with data augmentations
  provably isolates content from style. Advances in neural information
  processing systems  \textbf{34},  16451--16467 (2021)

\bibitem{laskin2020curl}
Laskin, M., Srinivas, A., Abbeel, P.: Curl: Contrastive unsupervised
  representations for reinforcement learning. In: International Conference on
  Machine Learning. pp. 5639--5650. PMLR (2020)

\bibitem{ba2016layer}
{Lei Ba}, J., {Kiros}, J.R., {Hinton}, G.E.: {Layer Normalization}. arXiv
  e-prints arXiv:1607.06450 (Jul 2016)

\bibitem{liu2021self}
Liu, X., Zhang, F., Hou, Z., Mian, L., Wang, Z., Zhang, J., Tang, J.:
  Self-supervised learning: Generative or contrastive. IEEE Transactions on
  Knowledge and Data Engineering  (2021)

\bibitem{lomanco2017core50}
Lomanco, V., Maltoni, D.: Core50: a new dataset and benchmark for continual
  object recognition. In: Proceedings of the 1st Annual Conference on Robot
  Learning. pp. 17--26 (2017)

\bibitem{marcel2010torchvision}
Marcel, S., Rodriguez, Y.: Torchvision the machine-vision package of torch. In:
  Proceedings of the 18th ACM international conference on Multimedia. pp.
  1485--1488 (2010)

\bibitem{mitrovic2020representation}
{Mitrovic}, J., {McWilliams}, B., {Walker}, J., {Buesing}, L., {Blundell}, C.:
  {Representation Learning via Invariant Causal Mechanisms}. arXiv e-prints
  arXiv:2010.07922 (Oct 2020)

\bibitem{murase2020can}
Murase, R., Suganuma, M., Okatani, T.: {How can CNNs use image position for
  segmentation?} arXiv e-prints arXiv:2005.03463 (May 2020)

\bibitem{nair2010rectified}
Nair, V., Hinton, G.E.: Rectified linear units improve restricted boltzmann
  machines. In: F{\"{u}}rnkranz, J., Joachims, T. (eds.) Proceedings of the
  27th International Conference on Machine Learning (ICML-10), June 21-24,
  2010, Haifa, Israel. pp. 807--814. Omnipress (2010)

\bibitem{nesterov1983amf}
Nesterov, Y.: A method for solving the convex programming problem with
  convergence rate $o(1/k^2)$. Proceedings of the USSR Academy of Sciences
  \textbf{269},  543--547 (1983)

\bibitem{radford2021learning}
Radford, A., Kim, J.W., Hallacy, C., Ramesh, A., Goh, G., Agarwal, S., Sastry,
  G., Askell, A., Mishkin, P., Clark, J., et~al.: Learning transferable visual
  models from natural language supervision. In: International Conference on
  Machine Learning. pp. 8748--8763. PMLR (2021)

\bibitem{rizve2021exploring}
Rizve, M.N., Khan, S., Khan, F.S., Shah, M.: Exploring complementary strengths
  of invariant and equivariant representations for few-shot learning. In:
  Proceedings of the IEEE/CVF Conference on Computer Vision and Pattern
  Recognition. pp. 10836--10846 (2021)

\bibitem{rumelhart1985learning}
Rumelhart, D.E., Hinton, G.E., Williams, R.J.: Learning internal
  representations by error propagation. Tech. rep., California Univ San Diego
  La Jolla Inst for Cognitive Science (1985)

\bibitem{schneider2021contrastive}
Schneider, F., Xu, X., Ernst, M.R., Yu, Z., Triesch, J.: Contrastive learning
  through time. In: SVRHM 2021 Workshop @ NeurIPS. pp. 1--14 (2021)

\bibitem{srinivas2019full}
Srinivas, S., Fleuret, F.: Full-gradient representation for neural network
  visualization. Advances in neural information processing systems  \textbf{32}
  (2019)

\bibitem{sutskever2013importance}
Sutskever, I., Martens, J., Dahl, G., Hinton, G.: On the importance of
  initialization and momentum in deep learning. In: International conference on
  machine learning. pp. 1139--1147. PMLR (2013)

\bibitem{tadepalli_english_2021}
Tadepalli, P.: English: {A} {Western} {Bluebird} sitting on a branch at in
  {Los} {Gatos}, {California}. (Mar 2021)

\bibitem{oord2018representation}
{van den Oord}, A., {Li}, Y., {Vinyals}, O.: {Representation Learning with
  Contrastive Predictive Coding}. arXiv e-prints arXiv:1807.03748 (Jul 2018)

\bibitem{wang2021residual}
Wang, Y., Geng, Z., Jiang, F., Li, C., Wang, Y., Yang, J., Lin, Z.: Residual
  relaxation for multi-view representation learning. Advances in Neural
  Information Processing Systems  \textbf{34},  12104--12115 (2021)

\bibitem{wang2021understanding}
Wang, Y., Huang, H., Rudin, C., Shaposhnik, Y.: Understanding how dimension
  reduction tools work: An empirical approach to deciphering t-sne, umap,
  trimap, and pacmap for data visualization. J. Mach. Learn. Res.
  \textbf{22}(201),  1--73 (2021)

\bibitem{xiao2020should}
{Xiao}, T., {Wang}, X., {Efros}, A.A., {Darrell}, T.: {What Should Not Be
  Contrastive in Contrastive Learning}. arXiv e-prints arXiv:2008.05659 (Aug
  2020)

\bibitem{zhang2020equivariance}
Zhang, L.: Equivariance and Invariance for Robust Unsupervised and
  Semi-Supervised Learning. Ph.D. thesis, University of Central Florida (2020)

\bibitem{zhang2019aet}
Zhang, L., Qi, G.J., Wang, L., Luo, J.: Aet vs. aed: Unsupervised
  representation learning by auto-encoding transformations rather than data.
  In: Proceedings of the IEEE/CVF Conference on Computer Vision and Pattern
  Recognition. pp. 2547--2555 (2019)

\end{thebibliography}
\end{document}